\documentclass[runningheads]{llncs}
\usepackage{graphicx}
\usepackage{amssymb}
\usepackage{amsmath}
\usepackage{color}
\usepackage{tabularray}
\definecolor{Silver}{rgb}{0.95,0.95,0.95}
\usepackage{soul}
\newcommand{\repeatthanks}{\textsuperscript{\thefootnote}}
\usepackage[
colorlinks=true,
urlcolor=blue,
linkcolor=black,
citecolor=black
]{hyperref}
\begin{document}
\title{Continual Multiple Instance Learning for Hematologic Disease Diagnosis}

\author{Zahra Ebrahimi\inst{1,2,3} \and
Raheleh Salehi \inst{1,4} \and
Nassir Navab \inst{5} \and \\
Carsten Marr \inst{1}\thanks{Co-corresponding: \{carsten.marr,ario.sadafi\}@helmholtz-munich.de} \and
Ario Sadafi \inst{1,5}\repeatthanks
}
% index{Ebrahimi, Zahra}
% index{Salehi, Raheleh}
% index{Navab, Nassir}
% index{Marr, Carsten}
% index{Sadafi, Ario}

\authorrunning{Z. Ebrahimi et al.}
\institute{Institute of AI for Health, Helmholtz Munich, Neuherberg, Germany \and 
TUM School of Computation, Information and Technology, Technical University Munich, Munich, Germany \and 
Faculty of Mathematics, Computer Science and Statistics
Ludwig-Maximilians-Universität München (LMU), Munich, Germany \and
Institute of Chemical Epigenetics, Faculty of Chemistry and Pharmacy,  Ludwig-Maximilians-Universität München (LMU), Munich, Germany \and
Computer Aided Medical Procedures, Technical University of Munich, Munich, Germany}

% \author{Anonymous}
%
% \author{Zahra Ebrahimi, Ario Sadafi, Raheleh Salehi, Nassir Navab, Carsten Marr}
% \authorrunning{Anonymous}

% \institute{}
% \institute{Anonymous Organization\\\email{**@*****.***}}
% \author{Anonymized Authors}  %% Added for anonymized MICCAI 2025 submission
% \authorrunning{Anonymized Author et al.}
% \institute{Anonymized Affiliations \\
%     \email{email@anonymized.com}}
\maketitle

\begin{abstract}
The dynamic environment of laboratories and clinics, with streams of data arriving on a daily basis, requires regular updates of trained machine learning models for consistent performance. Continual learning is supposed to help train models without catastrophic forgetting. However, state-of-the-art methods are ineffective for multiple instance learning (MIL), which is often used in single-cell-based hematologic disease diagnosis (e.g., leukemia detection).
Here, we propose the first continual learning method tailored specifically to MIL. Our method is rehearsal-based over a selection of single instances from various bags. We use a combination of the instance attention score and distance from the bag mean and class mean vectors to carefully select which samples and instances to store in exemplary sets from previous tasks, preserving the diversity of the data. Using the real-world input of one month of data from a leukemia laboratory, we study the effectiveness of our approach in a class incremental scenario, comparing it to well-known continual learning methods. 
We show that our method considerably outperforms state-of-the-art methods, providing the first continual learning approach for MIL.
This enables the adaptation of models to shifting data distributions over time, such as those caused by changes in disease occurrence or underlying genetic alterations.

\end{abstract}

\section{Introduction}
Computer-aided diagnosis systems based on deep learning have achieved human-level performance in high-throughput analysis of large amounts of image data \cite{van2021deep,matek2019human,eckardt2022deep2,campanella2019clinical}.
Multiple instance learning (MIL) plays a central role in medical image analysis. In MIL, a bag of instances is collected from a patient, and a single diagnosis is associated with all of them. These instances can be single cells or histopathology tissue patches -- on their own, they are normally insufficient to diagnose a patient, while together they can provide enough information for a reliable clinical conclusion.

In recent years, many methods have been developed for the diagnosis of haematologic diseases. For instance, Eckardt et al. proposed a method for the identification of acute promyelocytic leukemia (APL) \cite{eckardt2022deep} and acute myeloid leukemia (AML) with NPM1 mutation \cite{eckardt2022deep2}, based on single white blood cells microscopically imaged in a bone marrow smear. For detection of APL, Sidhom et al. \cite{sidhom2021deep} suggest an ensemble approach including both single instance and multiple instance level analysis. 
Attention-based MIL architectures provide an additional attention score indicating the most important instances that were used for the prediction. Focusing on single red blood cells, Sadafi et al. \cite{sadafi2020attention} proposed an approach for the diagnosis of hereditary hemolytic anemias from peripheral blood. Similarly, Hehr et al. \cite{hehr2023explainable} employed an attention-based MIL framework to classify genetic subtypes of AML.

With the huge stream of data arriving in laboratories on a daily basis, with the consistent changes observed in the data \cite{salehi2022unsupervised} regular updating of models is an ongoing task. However, training neural networks on new datasets can lead to forgetting previously learned tasks, an issue called catastrophic forgetting \cite{kemker2018measuring,sadafi2023continual}. Continual learning methods mitigate this problem in three different ways: (i) Augmenting the architecture of the model so that it has different sub-networks for every task; (ii) Regularizing updates of the weights of the network such that important weights are not updated while less important neurons contribute to learning the new task, e.g. via elastic weight consolidation (EWC) \cite{kirkpatrick2017overcoming}; (iii) Rehearsal-based methods such as iCaRL \cite{rebuffi2017icarl} sample a few images from the previous tasks and augment the dataset with those, such that they minimize the model's forgetting. 
Derakhshani et al. \cite{derakhshani2022lifelonger} have recently benchmarked continual learning methods on several biomedical datasets, demonstrating the potential of rehearsal-based approaches such as iCaRL. However, these methods are not applicable to MIL architectures, since sampling a huge bag of instances from histopathology or cytology datasets is not practically feasible. To the best of our knowledge, so far, no mechanism of instance sampling exists for efficiently selecting only the most diagnostically relevant instances while discarding less informative ones, which often constitute the majority of a bag's content. This gap highlights the need for a continual learning approach specifically tailored to MIL, enabling models to adapt efficiently while preserving their diagnostic accuracy. Given that MIL has become a fundamental paradigm in computational pathology and cytology, where diagnoses rely on aggregating information from large sets of instances rather than single observations, developing an effective continual learning strategy for MIL is crucial to ensuring applicability in evolving clinical settings.

Recently, MICIL \cite{meseguer2024micil} introduced a multiple-instance class-incremental learning framework for skin cancer whole slide image analysis, highlighting the potential of integrating MIL with continual learning in medical imaging. However, MICIL primarily addresses large-scale whole slide images and relies on embedding-level distillation to retain WSI-level knowledge. In contrast, a major challenge in many diagnostic settings, including hematology and histopathology, lies in the efficient selection of diagnostically relevant instances from vast and often heterogeneous patient samples containing hundreds of instances. In hematology, for instance, a large fraction of single cells in a patient sample do not contribute to the final diagnosis. Efficient instance selection is therefore crucial to ensure that continual learning methods focus on the most relevant instances and utilize the rehearsal memory efficiently.

Here, we are proposing CoMIL, the first continual learning method specifically tailored to MIL architectures and datasets with single instance selection. Our suggested rehearsal-based continual learning method performs instance sampling based on the attention score and its distance from the class mean and bag mean vectors in the latent space, such that the characteristics of the bags and hallmark cells necessary for diagnosis are preserved. 
% We provide the source code on our GitHub repository.% (NEED WEBPAGE HERE, left out for camera ready)
Source code is available at \url{https://github.com/marrlab/contmil}

\section{Methodology}

% \subsection{Attention based multiple instance learning}
% Multiple instance learning (MIL) methods use a bag of instances as input $B=\{I_1,\dots,I_N\}$ to provide a single prediction $\hat{y}$ and a set of attention values $\alpha_i \in A$ for every instance: $\hat{y}, A = f_\mathrm{MIL}(B; \theta_t)$. Attention based MIL consists of three stages: (i) For every instance $I_i$ features are extracted $h_i = \psi(I_i)$. (ii) based on the features an attention value $\alpha_i$ is estimated for every instance 
% \begin{equation}
%     \alpha_i = \frac{\mathrm{exp}\{w^T\mathrm{tanh}(Vh_i^T)\}}{\sum_{j=1}^N \mathrm{exp}\{w^T\mathrm{tanh}(Vh_j^T)\}}.
% \end{equation}
% (iii) Based on the attention values, a bag feature vector is calculated $z = \sum^N_{k=1} \alpha_kh_k$ and used for the final classification.  
\subsection{Attention-based MIL}

\begin{figure}[t]
\centering
\includegraphics[width=\textwidth,page=1,trim=0cm 9.5cm 0cm 0cm,clip]{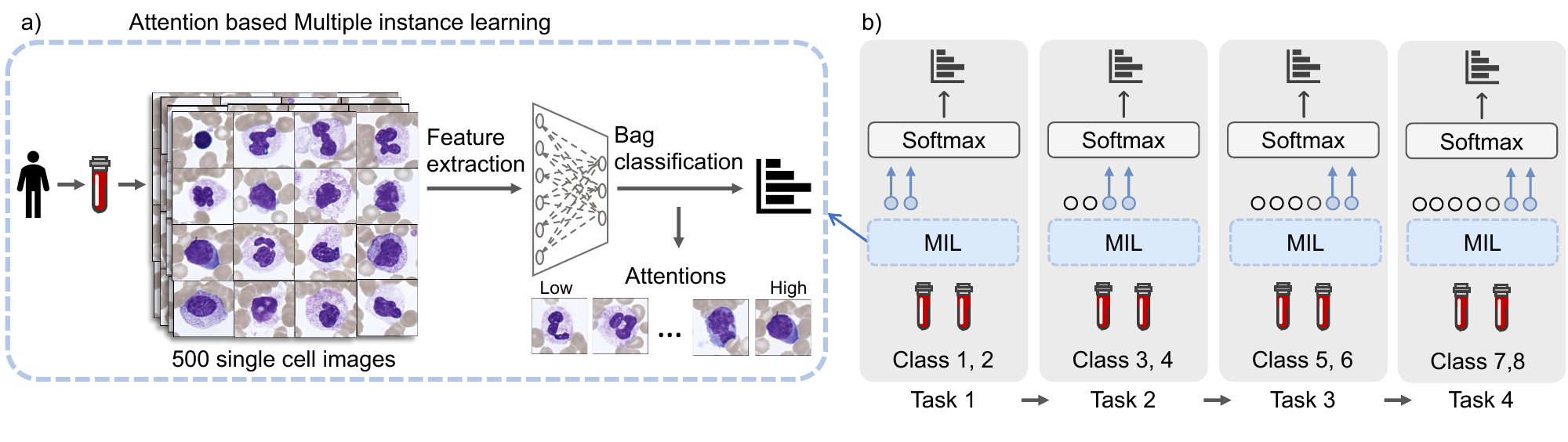}
\caption{A continual attention-based MIL scenario for hematologic disease diagnosis. a) From a blood probe, 500 single white blood cell images (instances) are taken for one patient (bag). After feature extraction and pooling, the bag classification probability and instance attentions are returned. b) Our dataset consists of 8 hematologic disease classes, and we have divided it into four tasks to evaluate our class incremental learning method.}
\label{figoverview}
\end{figure}

Multiple instance learning methods operate on a bag of instances $B = \{I_1, \dots, I_N\}$ to produce a single bag-level prediction $\hat{y}$ along with a set of attention scores $\alpha_i \in A$ for each instance. Attention-based MIL consists of three stages (see Fig. \ref{figoverview}):
% \[
% \hat{y}, A = f_\mathrm{MIL}(B; \theta_t),
% \]
% where $f_\mathrm{MIL}$ is the MIL model parameterized by $\theta_t$ at task $t$, and $N$ is the number of instances in the bag.

\begin{enumerate}
    \item \textbf{Instance feature extraction:} For each instance $I_i$, a feature vector $h_i = \psi(I_i)$ is computed, where $\psi$ is the pretrained feature extractor.\\
    
    \item \textbf{Attention weight computation:} The importance of each instance is estimated through an attention mechanism. Attention for instance $I_i$ is computed as:
    \begin{equation}
        \alpha_i = \frac{\exp\left\{ w^\top \tanh(V h_i^\top) \right\}}{\sum_{j=1}^{N} \exp\left\{ w^\top \tanh(V h_j^\top) \right\}},
    \end{equation}
    Where $V \in \mathbb{R}^{D \times d}$ is a trainable matrix that projects instance features $h_i \in \mathbb{R}^d$ into a $D$-dimensional attention space, and $w \in \mathbb{R}^{D}$ is a trainable vector that maps the transformed features to scalar attention scores.\\
    
    \item \textbf{Bag-level aggregation:} The bag feature vector $z$ is computed as a weighted sum of the instance features:
    \begin{equation}
    z = \sum_{k=1}^{N} \alpha_k h_k,
    \end{equation}
    Which aggregates information from the most relevant instances in the bag. This bag-level representation $z$ is then used for the final classification.
\end{enumerate}

\section{Evaluation}

\subsection{Problem definition}
At every step $t$ a training set $\mathcal{D}_t = \{B, y\}$ is provided such that $y_i \in C_t$ is a label from the set of new classes $C_t$ the classifier $f_t(.)$ is going to be trained on. The training involves updating model parameters $\theta_t$, having $\theta_{t-1}$ as the knowledge from the previous task. 
An exemplar set $\mathcal{X}_t$ is sampled at every step, containing bags with selected instances and stored in a limited amount of memory capable of holding $K$ instances. This set of samples is carried to the next step of continual learning and used for rehearsal of the previous knowledge during the training.

\subsection{Training}
At every step $t$, the training set $\mathcal{D}_t$ is augmented with the exemplar set $\mathcal{X}_t$. The model is initialized with the previously learned parameters $\theta_{t-1}$. The minimization is performed using a loss consisting of distillation and classification losses, defined as:

\begin{equation}
    \mathcal{L}(\theta_t) = \mathcal{L}_\mathrm{cls}(\theta_t)  + \mathcal{L}_\mathrm{dist}(\theta_t) .
\end{equation}
and a negative log likelihood loss for the classification loss. 
% \begin{equation}
%     \mathcal{L}_\mathrm{cls} = \sum_{j \in C_t} -y_j\mathrm{log}(\hat{y_j}).
% \end{equation}
The distillation loss preserves the previous state of the model as much as possible and alleviates catastrophic forgetting. Having $\theta_{t-1}$ as the model parameters before the beginning of training step $t$, distillation loss is defined as

\begin{equation}
    \mathcal{L}_{\mathrm{dist}}(\theta_t) =\sum_{j \in C_{t-1} }  \mathcal{S} (\hat{y}_j^{t-1})\mathrm{log}( \mathcal{S}(\hat{y}_j^{t})) + (1 - \mathcal{S} (\hat{y}_j^{t-1})) \mathrm{log}(1 - \mathcal{S}(\hat{y}_j^{t})) 
\end{equation}
where
$\forall (B_j) \in \mathcal{D}_t \cup \mathcal{X}_t$: $\hat{y}_j^{t-1} = f_\mathrm{MIL}(B_j; \theta_{t-1})$ and $\hat{y}_i^{t}=f_\mathrm{MIL}(B_j; \theta)$. $\mathcal{S}(.)$ is the sigmoid function and $C_{t-1}$ denotes the classes at $t-1$.

\subsection{Exemplar set selection}
To rehearse the previously learned classes, a small number of instances can be carried between the tasks, as commonly done in rehearsal-based continual learning \cite{bagus2021investigation}.
An exemplar set selection must be specifically tailored for MIL methods and carried out at the instance level. Using existing methods such as iCaRL \cite{rebuffi2017icarl} and keeping whole bags is memory-consuming and severely limits the number and diversity of the selected examples. 

With a limited memory of $K$ instances and $|C_t|$ classes, a total of $m =\frac{K}{|C_T|}$ instances can be selected for each class. 
Considering that not all instances are equal in information value, one intuitive way of sampling is to keep instances with the highest attention in each bag. However, this method fails to preserve the bag structure, which is formed mainly by non-diagnostic instances and breaks the structure of the bag, leading to a shift in attention pooling outcome and hence the final classification (see Sec. \ref{sec:results} and Fig. \ref{figresult}).

Having all of the instances in all bags of any class $I_j \in B_j \in \mathcal{D}_c$ where $\mathcal{D}_c$ are all of the samples in the dataset belonging to class $c$.
For every instance, the CoMIL instance value is defined as:

\begin{equation}
    v_i = \alpha_i + \sqrt{(\psi(I_i;\theta_t) - \Bar{B_j})^2} + \sqrt{(\psi(I_i;\theta_t) - \Bar{C})^2}
    \label{eq:val}
\end{equation}
where for every bag $B_j$ with a size of $N$, a bag mean is defined as $\Bar{B_j} = \frac{1}{N} \sum_{i=1}^N I_i$ and a class mean is also defined as $\Bar{C} = \frac{1}{M}\sum_{j=1}^M B_j$ where $M$ is the total number of bags in the class. $\alpha_i$ is the attention value the model has estimated for the instance.

After estimation of every CoMIL instance value, considering a memory capacity of $K$ instances and the fact that every instance occupies one unit of memory, a typical knapsack problem is formulated \cite{mathews1896partition}.
Here, the goal is to select items to fill the capacity while maximizing the amount of total value of the selected items. We decided to solve this problem with dynamic programming \cite{cormen01introduction}. After selection of instances, they are organized back into their corresponding bags and stored in the exemplar set.

\subsection{Exemplar set reduction}
With the arrival of new classes, the amount of memory associated with each class decreases, and shrinking the exemplar sets is inevitable. If the old size of the memory for every class is $m' = \frac{K}{{C_{t-1}}}$ and the new memory size is $m = \frac{K}{C_t}$, the value for all of the instances are once more estimated using equation (\ref{eq:val}). This time, we shrink the bags by removing the least important instances. If a bag is empty, it will be completely removed from exemplar sets.

\begin{table}
\centering
% \caption{Classes in the dataset with number of patients in training and test set. The task number indicates at which stage of our simulated continual learning task the model is trained on the respective class.}
\caption{Classes with training and test patient counts; the task number indicates when each class is introduced in the simluated continual learning process.}
\label{tab:cls}
\begin{tblr}{
  width = \linewidth,
  colspec = {Q[543]Q[123]Q[122]Q[100]},
  column{even} = {c},
  column{3} = {c},
  vline{2-4} = {-}{},
  hline{2-9} = {-}{},
}
 & Training patients & Test patients & Task number\\
No disease evidence & 348 & 116 & 1\\
Acute myeloid leukemia (AML) & 34 & 12 & 1\\
Myelodysplastic syndrome (MDS) & 133 & 44 & 2\\
Chronic myeloid leukaemia (CML) & 46 & 16 & 2\\
Chronic lymphocytic leukemia (CLL) & 91 & 30 & 3\\
Myeloproliferative neoplasms (MPN) & 192 & 64 & 3\\
Chronic myelomonocytic leukemia (CMML) & 37 & 13 & 4\\
Lymphoma (LYM) & 34 & 11 & 4
\end{tblr}
\end{table}

\subsection{Dataset and continual learning scenario}
We are testing our method on a dataset with 1221 patient samples, each containing 500 microscopy single white cell images, resulting in more than 611,000 single cell images. There are 8 classes in the dataset, comprising patients with and without neoplastic changes (Table \ref{tab:cls}). In our class incremental experiment, we have defined 4 steps, where in each step, 2 new classes are introduced. We have divided the data into 75\% training and 25\% hold-out test sets in a stratified manner. 
Figure \ref{figoverview} shows the continual learning scenario and MIL architecture.

\subsection{Baselines}
    To evaluate the performance of our continual learning method, we compare it against six different baselines: (i) Simple finetuning, a naive approach to the problem; (ii) Elastic Weight Consolidation (EWC) \cite{kirkpatrick2017overcoming}, a regularization-based solution for continual learning; (iii) iCaRL \cite{rebuffi2017icarl}, an established rehearsal-based continual learning method; (iv) BIC \cite{wu2019large}, a bias correction method designed to mitigate the imbalance between old and new classes in incremental learning; (v) DER \cite{yan2021dynamically}, a memory-efficient continual learning approach that combines knowledge distillation with replay strategies; and (vi) attention-based iCaRL, where instances are treated according to their attention value.

\subsection{Implementation details}
Our MIL architecture consists of two networks. The first network is a ResNext \cite{xie2017aggregated}, pretrained on the AML cytomorphology dataset \cite{matek2019single,clark2013cancer} for the task of single leukocyte recognition \cite{matek2019human}, with its parameters frozen and used as a feature extractor. 
The MIL network has three layers of convolutions followed by a two-layer fully connected classifier. The attention module also has two fully connected layers.
In our experiments, the model is trained for 20 epochs using a stochastic gradient descent optimizer. 
For iCaRL and all rehearsal-based experiments, the exemplar set size $K$ is set to $5000$.

\section{Results}
\label{sec:results}
\begin{figure}[t]
\centering
\includegraphics[width=0.80\textwidth,page=2,trim=0cm 6cm 7cm 0cm,clip]{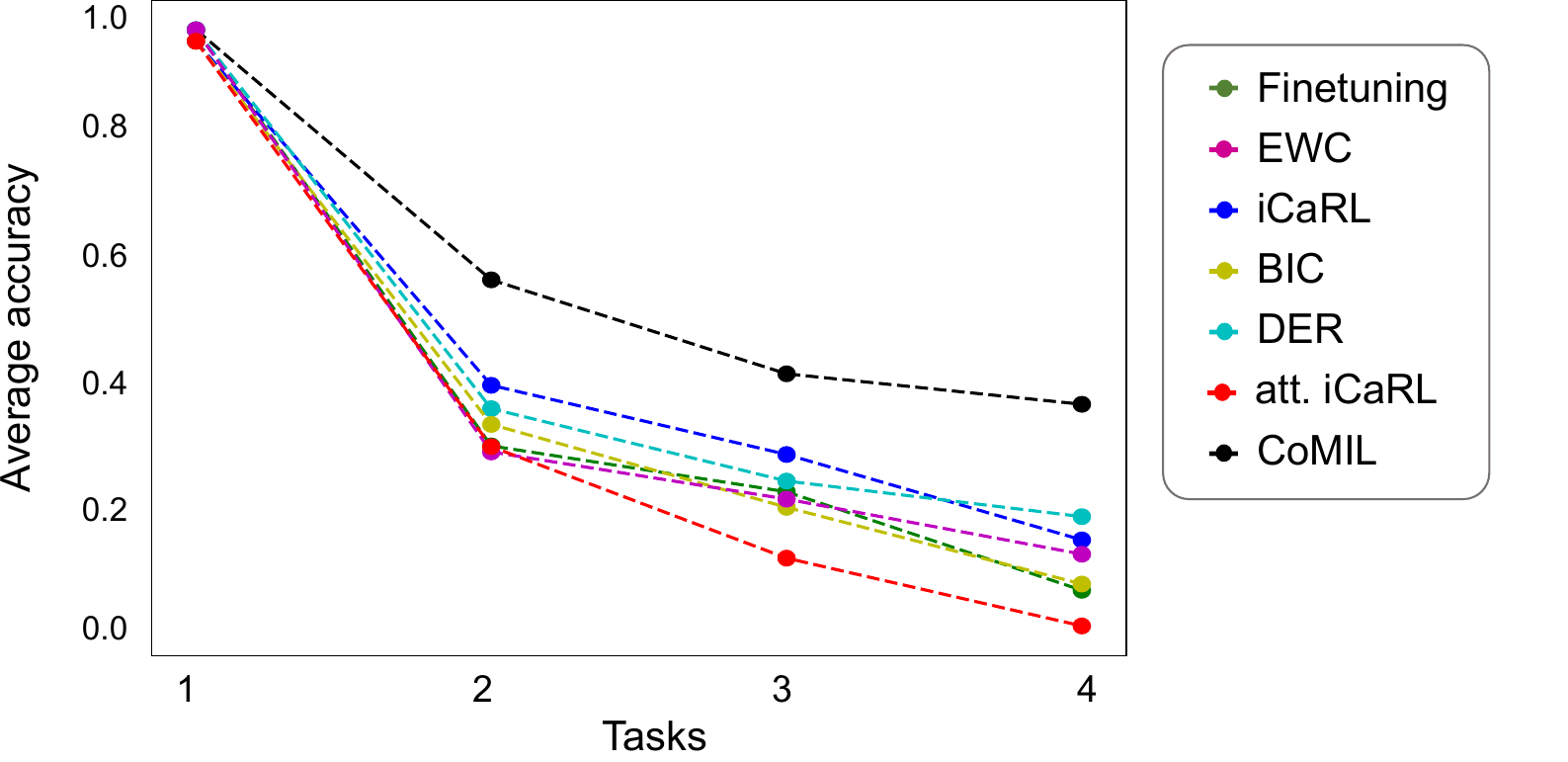}
\caption{Average accuracy of CoMIL on four steps of class increment learning experiment surpasses simple finetuning, elastic weight consolidation (EWC), iCaRL, Bias Correction (BIC), Dynamically Expandable Representation (DER), and iCaRL exemplar set sampling based on high attention cells of every bag (Attention iCaRL).}
\label{figresult}
\end{figure}

In addition to the baselines, we conducted a comparison between our proposed method with an upper bound where the model has access to the complete datasets from previous tasks. The results are summarized in Table \ref{tab:res}, which presents the average accuracy and average forgetting \cite{chaudhry2018riemannian} of our CoMIL approach, as well as the baselines and the defined upper bound (UB). CoMIL consistently outperforms the baseline methods, indicating its superiority in handling multiple instance learning tasks.
Figure \ref{figresult} shows the average accuracy for four tasks defined in our class incremental learning experiment. 

% % \usepackage{tabularray}
% \begin{table}
% \centering
% \caption{Average accuracy and average forgetting is reported for all methods on all four tasks and classes within each task. CoMIL method consistanly outperforms other baselines of finetuning, elastic weight consolidation (EWC), iCaRL, and attention based iCarl (Att. iCaRL). Upper bound (UB) of the problem is also when the model has access to all previous data.}
% \label{tab:res}
% \begin{tblr}{
%   width = \linewidth,
%   colspec = {Q[70]Q[130]Q[130]Q[130]Q[130]Q[130]Q[130]Q[130]Q[110]},
%   column{even} = {c},
%   column{3} = {c},
%   column{5} = {c},
%   column{7} = {c},
%   column{9} = {Silver,c},
%   vline{2-8} = {-}{},
%   hline{2-3} = {-}{},
% }
%  & Fine- tuning & EWC & iCaRL & BIC & DER & att. iCarl & CoMIL & UB  \\
% Acc.$\uparrow$ & 0.45\tiny±0.01 & 0.46\tiny±0.01 & 0.50\tiny±0.01 & 0.48\tiny±0.02 & 0.46\tiny±0.02&0.42\tiny±\hl{0.xx}&0.61\tiny±0.01 & 0.70\tiny±0.15\\
% For.$\downarrow$ & 0.32\tiny±0.01 & 0.76\tiny±0.01 & 0.24\tiny±0.02 & 0.26\tiny±0.01 & 0.26\tiny±0.01 & 0.30\tiny±\hl{0.xx} & 0.26\tiny±0.02 &-
% \end{tblr}
% \end{table}

\begin{figure}[t]
\centering
\includegraphics[width=0.70\textwidth,page=3,trim=0cm 11cm 11cm 0cm,clip]{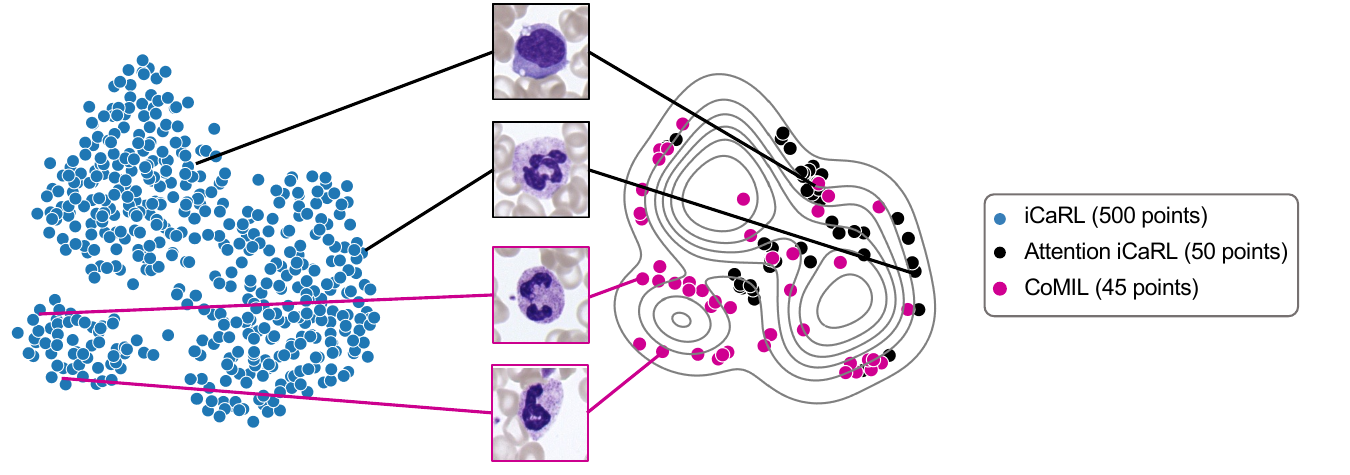}

\caption{CoMIL sampling leads to a better preservation of the bag structure for continual learning as compared to other methods. Comparison of sampling between three different methods on the instance level within a bag is displayed. iCaRL selects all instances, while attention-based iCaRL only keeps high-attention instances. A better distribution of sampling points is only done through our proposed approach.}
\label{figumap}
\end{figure}

To gain a better understanding of the instance selection process within the bags, we plotted a UMAP embedding of the selected instances in Figure \ref{figumap}. In the case of iCaRL, which lacks an instance selection mechanism, all 500 images in the bag are sampled (blue dots in Fig. \ref{figumap} left). When using an attention-based selection method for iCaRL, only the 50 most important instances are selected (black dots, right). However, this selection is limited to diagnostic hallmark cells, and the original structure of the bag is lost, resulting in a shift in the bag feature vector. Visually, rarely any point covers the third maximum in the KDE plot, probably containing mainly granulocytes (Fig. \ref{figumap} middle images with pink frame).

\begin{table}
\centering
\caption{Average accuracy and average forgetting are reported for all methods on all four tasks and classes within each task. CoMIL method consistently outperforms other baselines of finetuning, elastic weight consolidation (EWC), iCaRL, Bias Correction (BIC), Dynamically Expandable Representation (DER), and attention-based iCaRL (att. iCaRL). The upper bound (UB) of problem assumes model access to all previous data.}
\label{tab:res}
\begin{tblr}{
  width = \linewidth,
  colspec = {Q[90]Q[140]Q[140]Q[140]Q[140]Q[140]Q[140]Q[140]Q[140]},
  column{even} = {c},
  column{3} = {c},
  column{5} = {c},
  column{7} = {c},
  column{9} = {Silver,c},
  vline{2-8} = {-}{},
  hline{2-3} = {-}{},
}
 & Fine- tuning & EWC & iCaRL & BIC & DER & att. iCarl & CoMIL & UB  \\
Acc.$\uparrow$ & 45.2\tiny±0.1 & 46.0\tiny±0.1 & 49.9\tiny±0.1 & 45.3\tiny±0.2 & 49.4\tiny±0.2&43.1\tiny±0.3&\textbf{61.3}\tiny±0.1 & 70.0\tiny±15.0\\
For.$\downarrow$ & 31.5\tiny±0.1 & 75.5\tiny±0.1 & 24.2\tiny±0.2 & 26.5\tiny±0.1 & 25.0\tiny±0.1 & 31.8\tiny±0.3 & \textbf{22.1}\tiny±0.2 &-
\end{tblr}
\end{table}
Our proposed CoMIL instance selection method achieves a more uniform distribution of points, covering all of the sub-distributions in the KDE plot (pink dots). This emphasizes the effectiveness of our approach in preserving the diversity of the data and selecting representative instances for continual learning in MIL tasks.

To mitigate privacy concerns in medical data replay, our method can operate solely on pre-extracted, anonymized feature embeddings rather than raw images or identifiable patient data. The exemplar memory stores only these anonymized features, further minimizing re-identification risk. Additionally, patient data is excluded from single-cell images.

\section{Conclusion}
We proposed CoMIL, an approach to continual multiple instance learning, and showed its potential on a challenging hematologic disease diagnosis task. The proposed method is rehearsal-based and uses instance attention scores and distances from bag mean and class mean vectors as criteria to select samples and instances to store in exemplary sets from previous tasks. 

We chose to use the knapsack method for instance sampling to provide a more versatile and adaptable approach for future developments, rather than relying on a simpler algorithm like the greedy method, which would only work for instances with equal cost. Our method allows for a range of possibilities, such as incorporating instances with varying costs, including higher costs for those that initiate a new bag, or adjusting costs to balance the selection of instances based on the difficulty or prevalence of a particular class. There are many exciting directions for future enhancements of our approach.

We evaluated our approach using real-world data from a leukemia laboratory, comprising over 600,000 single-cell images. In a class-incremental setting, CoMIL significantly outperformed established continual learning methods, demonstrating the effectiveness of the first continual learning approach tailored for MIL. This study marks an important step toward developing more efficient and accurate machine learning models for disease diagnosis in clinical and laboratory environments, where new data arrives daily and regular model updates are essential, making continual learning not just beneficial but necessary.

\subsubsection*{Acknowledgement} C.M. acknowledges funding from the European Research Council (ERC) under the European Union's Horizon 2020 research and innovation program (Grant Agreement No. 866411 \& 101113551 \& 101213822) and support from the Hightech Agenda Bayern.

\subsubsection*{Disclosure of Interests}
The authors have no competing interests to declare that are relevant to the content of this article.

\bibliographystyle{splncs04}
\bibliography{article}
\end{document}